\documentclass[final,3p,times]{elsarticle}

\usepackage{amssymb}
\usepackage{pifont}
\usepackage{longtable}
\usepackage{amsmath}
\usepackage{colortbl}
\usepackage{booktabs}
\usepackage{caption}
\usepackage{xcolor}
\usepackage{float}
\usepackage{bm}
\usepackage{graphicx}
\usepackage{amssymb}
\usepackage{orcidlink}
\usepackage{algorithm}
\usepackage{algpseudocode}
\usepackage{multirow}
\usepackage{colortbl}
\usepackage{graphicx}

\usepackage{longtable}
\usepackage{booktabs}
\usepackage{amsthm}
\usepackage{supertabular}

\definecolor{top1}{RGB}{255,0,0} 
\definecolor{top2}{RGB}{0,0,255} 
\definecolor{top3}{RGB}{0,255,0} 
\definecolor{top4}{RGB}{128,0,128} 

\newcommand{\del}[1]{\iffalse{#1}\fi}

\newcommand{\rem}[1]{\iffalse{#1}\fi}

\journal{Information Fusion}

\begin{document}

\begin{frontmatter}



\title{TSJNet: A Multi-modality Target and Semantic Awareness Joint-driven Image Fusion Network}


\author[author1]{Yuchan Jie \fnref{co-first}}
\ead{jyc981214@163.com}

\author[author1]{Yushen Xu \fnref{co-first}}
\ead{2112355010@stu.fosu.edu.cn}

\author[author1]{Xiaosong Li\corref{cor}}
\ead{lixiaosong@buaa.edu.cn}

\author[author2]{Huafeng Li}
\ead{hfchina99@163.com}

\author[author1]{Haishu Tan}
\ead{tanhaishu@fosu.edu.cn}

\author[author3]{Feiping Nie}
\ead{feipingnie@gmail.com}

\fntext[co-first]{Equal Contribution}
\cortext[cor]{Corresponding author}

\address[author1]{School of Physics and Optoelectronic Engineering, Foshan University, 528225, Foshan, China}

\address[author2]{ School of  Information Engineering and Automation, Kunming University of Science and Technology, 650500, Kunming, China}

\address[author3]{School of Artificial Intelligence, Optics and Electronics (i0PEN), School of Computer Science, Northwestern Polytechnical University, Xi'an 710072, China}



\begin{abstract}
This study aims to address the problem of incomplete information in unimodal images for semantic segmentation and object detection tasks. Existing multimodal fusion methods suffer from limited capability in discriminative modeling of multi-scale semantic structures and salient target regions, which further restricts the effective fusion of task-related semantic details and target information across modalities. To tackle these challenges, this paper proposes a novel fusion network termed TSJNet, which leverages the semantic information output by high-level tasks in a joint manner to guide the fusion process. Specifically, we design a multi-dimensional feature extraction module with dual parallel branches to capture multi-scale and salient features. Meanwhile, a data-agnostic spatial attention module embedded in the decoder dynamically calibrates attention allocation across different data domains, significantly enhancing the model's generalization ability. To optimize both fusion and advanced visual tasks, we balance performance by combining fusion loss with semantic losses. Additionally, we have developed a multimodal unmanned aerial vehicle (UAV) dataset covering multiple scenarios (UMS). Extensive experiments demonstrate that TSJNet achieves outstanding performance on five public datasets (MSRS, M\textsuperscript{3}FD, RoadScene, LLVIP, and TNO) and our UMS dataset. The generated fusion results exhibit favorable visual effects, and compared to state-of-the-art methods, the mean average precision (mAP@0.5) and mean intersection over union (mIoU) for object detection and segmentation, respectively, improve by 7.97\% and 10.88\%.The code and the dataset has been publicly released at \url{https://github.com/XylonXu01/TSJNet}.
\end{abstract}



\begin{keyword}
Multi-modal image fusion, object detection, semantic segmentation, benchmark.


\end{keyword}

\end{frontmatter}




\section{Introduction}
Multisensor integration is vital for intelligent technologies, such as unmanned aerial vehicle (UAV) precision combat and autonomous driving. As sensor technology advances, leveraging multi-modal images becomes crucial for complex real-world scene analysis. Visible and infrared sensors, commonly employed, exhibit limitations. While visible light sensors struggle to highlight targets effectively under low-light conditions, infrared sensors, unaffected by this issue, provide low scene resolution and capture texture details poorly \cite{13}. Fortunately, multi-modal image fusion (MMIF) can mitigate these challenges by synthesizing data from multiple sources, compensating for information loss in single- modal data, and enhancing understanding and perception for both human and machine vision. Moreover, MMIF is anticipated to yield more accurate representations of targets and scenes, positively impacting downstream tasks such as semantic segmentation and object detection \cite{sun2022detfusion, 23}.
\begin{figure}[ht]
  \centering
   \includegraphics[width=1.0\linewidth]{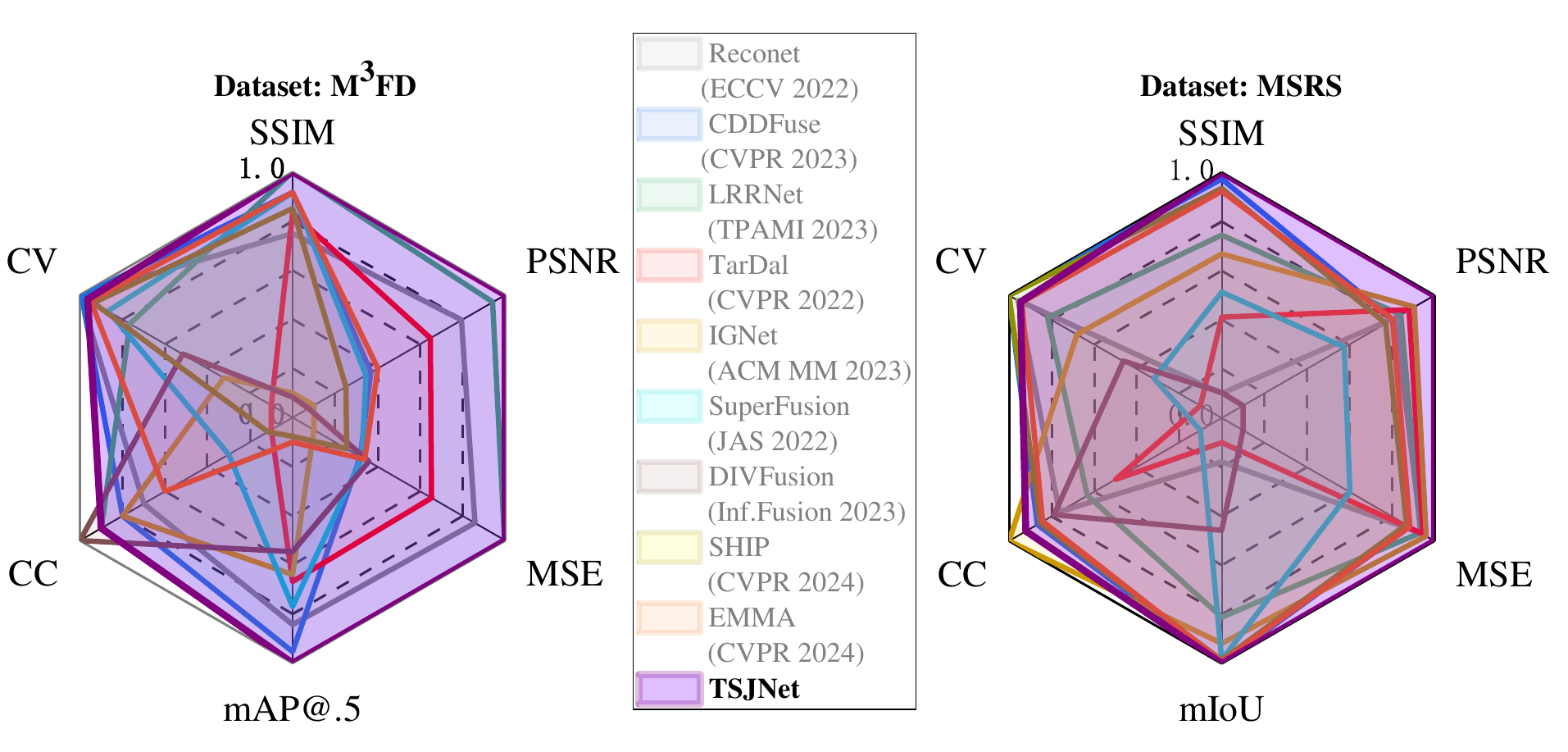}
   \caption{Comparison of results with SOTA methods on the M\textsuperscript{3}FD \cite{Liu_Fan_Huang_Wu_Liu_Zhong_Luo} and MSRS \cite{Tang_Yuan_Zhang_Jiang_Ma} datasets. The radar map highlights the superiority of the TSJNet.}
   \label{fig1}
\end{figure}

Recent advancements in deep learning (DL) feature extraction have propelled the development of DL-based MMIF methods. Notably, convolutional neural networks (CNN) \cite{02}, generative adversarial networks (GAN) \cite{Liu_Fan_Huang_Wu_Liu_Zhong_Luo}, transformer\cite{08}, diffusion models \cite{03, 04} and autoencoders (AE) \cite{Li_Xu_Wu_Lu_Kittler_2023, Zhao_Bai_Zhang_Zhang_Xu_Lin_Timofte_Gool_2022} have emerged as key approaches in this field, owing to their robust capabilities in feature representation. However, these approaches encounter two problems. First, existing multimodal fusion methods suffer from deficiencies in multi-scale semantic modeling and salient target feature characterization, which renders them unable to effectively integrate cross-modal semantic details and target information. Second, existing methods overlook the synergistic relationship between fusion tasks and subsequent high-level tasks. Third, existing multimodal image fusion datasets are predominantly derived from ground-level perspectives, lacking multi-scenario and multi-condition samples that cover UAV aerial photography. This deficiency restricts the generalization capability of current methods in complex aerial environments.

To address the first issue, existing fusion models \cite{Zhang_Jiao_Ma_Liu_Liu_Li_Chen_Yang_2023,05,11} still suffer from deficiencies in multi-scale semantic modeling and salient target feature characterization, which renders them unable to effectively integrate cross-modal task-relevant semantic details and target information. Taking infrared-visible image fusion (IVIF) as a typical example, although the two modalities are captured from the same target scene, their feature representations exhibit substantial discrepancies: the large-scale contours of targets belong to shared features, whereas the thermal radiation information in infrared images and the texture edge information in visible-light images are the salient features specific to each modality, respectively. Despite the Dif-Fusion model \cite{Yue_Fang_Xia_Deng_Ma_2023} integrating infrared and visible-light images into multi-channel inputs and extracting fused features via a diffusion model, it lacks a dedicated mechanism tailored to the disparities between shared and salient features. Consequently, it remains difficult for the model to fully integrate global and local information across multiple modalities. It is thus evident that there is an urgent need for an algorithm capable of simultaneously capturing shared and salient features, as well as achieving effective fusion of multi-scale semantic and target information.

For the second problem, a valuable MMIF model produces high-quality images in real-world scenarios and supports downstream tasks. While some existing methods \cite{Li_Chen_Liu_Ma_2023,22, 14,15} demonstrate good fusion performance, they often neglect to achieve mutual benefits between downstream tasks and image fusion. For example,
CDDFuse \cite{Zhao_Bai_Zhang_Zhang_Xu_Lin_Timofte_Gool_2022} incorporates an autoencoder based on Restomer and a feature fusion layer based on Lite Transformer and reversible neural networks. While this approach takes complementary feature extraction into account during the fusion process, it places insufficient emphasis on semantic information. Fortunately, recent studies \cite{Zhao_Xie_Zhao_Lu} have established a strong connection between image fusion and object detection or semantic segmentation, though they may not fully explore the deeper correlations among these three aspects.

Additionally, in military security and combat scenarios, aerial imagery presents challenges for drone-based object detection and semantic segmentation due to complex backgrounds and varying lighting conditions. However, existing MMIF datasets \cite{ toet2017tno, Xu_Ma_Le_Jiang_Guo_2020, Jia_Zhu_Li_Tang_Zhou_2021, Tang_Yuan_Zhang_Jiang_Ma, Liu_Fan_Huang_Wu_Liu_Zhong_Luo} primarily focus on traditional ground-level perspectives, significantly limiting the adaptability of current fusion methods to the diverse environments encountered from a drone's viewpoint, thereby impacting their effectiveness in practical applications.

\begin{figure}[htbp]
\centering	
\includegraphics[width=0.6\linewidth]{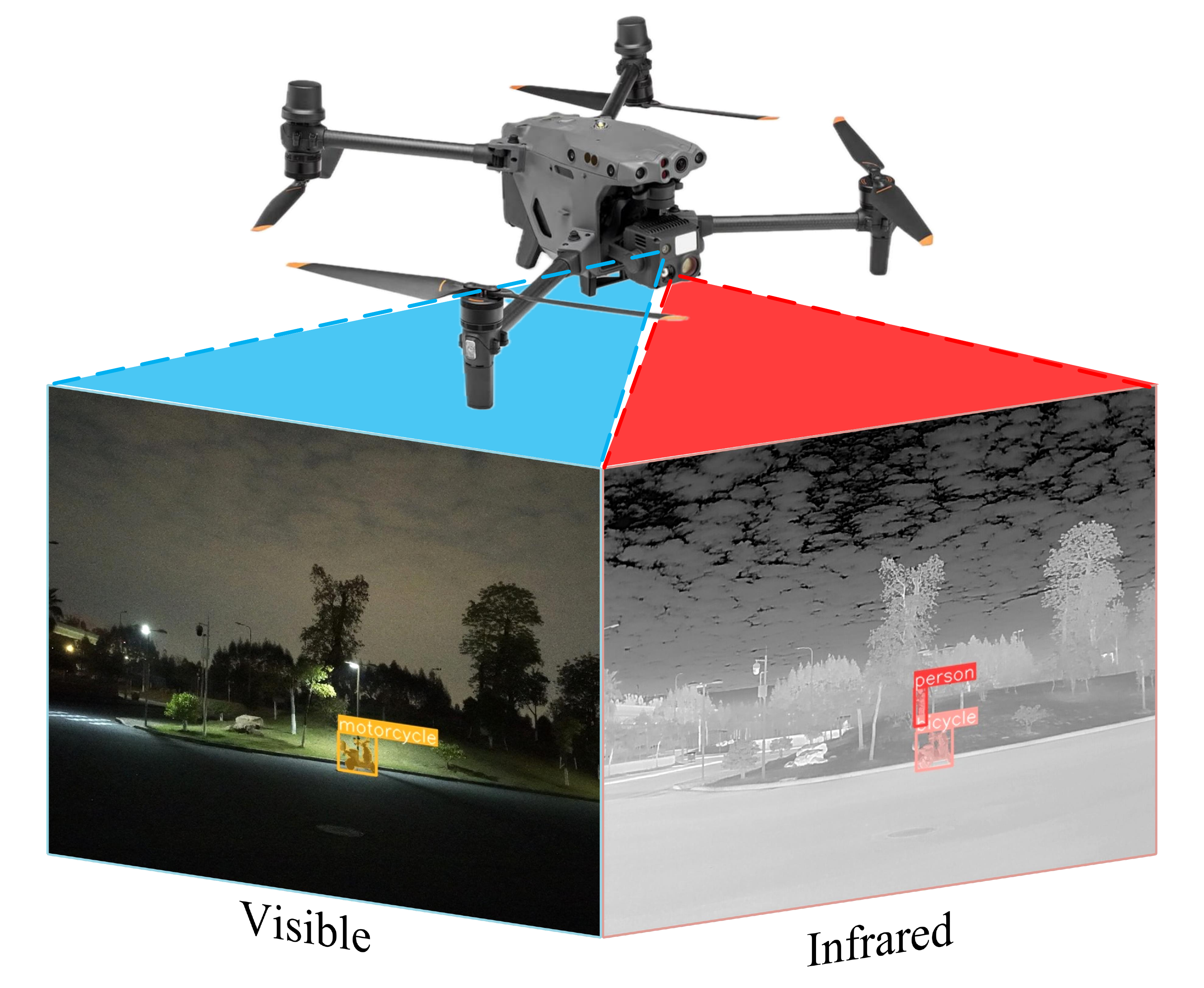} 
\caption{Unmanned Aerial Vehicle Data Acquisition System.The visible and infrared image pairs in the same scene are captured using a UAV equipped with both a zoom camera and an infrared camera.}
\label{fig.8}
\end{figure}

To address the aforementioned challenges, we propose TSJNet, a multi-feature learning MMIF network designed to effectively integrate multi-scale shared features and salient modality-specific information for downstream segmentation and detection tasks. 
Specifically, we design a Multi-Dimensional Feature Extraction Module (MDM) with a dual-branch structure that captures complementary multi-scale semantic features and salient modality-specific features, enabling effective integration of cross-modal semantic details and target information.
Additionally, we integrate semantic segmentation and object detection networks to guide the refinement of fusion features. This is achieved by combining segmentation and detection losses with the fusion loss, enhancing the accuracy of fusion results in downstream tasks. Fig.~\ref{fig1} demonstrates that the proposed network outperforms nine state-of-the-art (SOTA) methods in fusion performance. Additionally, Fig.~\ref{fig.8} illustrates our construction of a new benchmark for the UAV multi-scenario dataset (UMS), encompassing both aerial and conventional viewpoints. Our contributions are as follows:

\begin{enumerate}
\item We present TSJNet, an MMIF network that synergistically combines detection and segmentation. TSJNet generates high-quality fusion images and significantly improves the accuracy of unimodal images in object detection and semantic segmentation tasks.

\item We design the MDM with dual parallel branches to capture multi-scale and salient features, facilitating comprehensive cross-modal representation. 
    
\item  We create a comprehensive multi-modal UAV dataset consisting of 346 registered image pairs across six diverse scenarios\del{, including annotations for six target types}, suitable for image fusion, detection and segmentation tasks.

\end{enumerate}

\section{Related work}

\subsection{Multi-modal image fusion}
Over recent years, numerous DL-based MMIF approaches have emerged\cite{18, 10}, primarily falling into five categories: CNN- \cite{Liu_Liu_Wu_Ma_Liu_Zhong_Luo_Fan}, GAN- \cite{Ma_Zhang_Shao_Liang_Xu_2021}, transformer-\cite{li2026all,08,21}, diffusion model- \cite{Zhao_Bai_Zhu_Zhang_Xu_Zhang_Zhang_Meng_Timofte_Gool_2023,Yue_Fang_Xia_Deng_Ma_2023,20}, and AE-\cite{Zhao_Bai_Zhang_Zhang_Xu_Lin_Timofte_Gool_2022,Qu_Liu_Wang_Song_2021,Li_Wu_Kittler_2021,li2025umcfuse} based methods. 

\begin{table*}[]
\centering
\caption{Comparison of the proposed UMS dataset with five commonly used public multi-modal datasets}
\label{table1}
\resizebox{\textwidth}{!}{%
\begin{tabular}{ccccccc}
\hline
Datasets  & Image Pairs & Resolution & Camera angle          & Poor light & Objects  & Annotation                 \\ \hline
TNO       & 261         & 768×576    & ground                & few        & few      & ×                          \\
RoadScene & 221         & 768×576    & drive                 & moderate   & moderate & ×                          \\
LLVIP     & 16836       & 1080×720   & surveillance          & \checkmark          & \checkmark        & ×                          \\
MSRS      & 1444        & 640×480    & drive                 & moderate   & \checkmark        & Segmentation               \\
M\textsuperscript{3}FD      & 4200        & 1024×768   & ground, drive         & moderate   & \checkmark        & Detection                  \\
UMS       & 346         & 640×512    & \textbf{ground, drive, flight} & \textbf{\checkmark}          & \textbf{\checkmark}        & \textbf{Segmentation, Detection} \\ \hline
\end{tabular}%
}
\vspace{-10pt}
\end{table*}

Despite remarkable progress in DL for low-level vision\rem{ tasks}, driven by the nonlinear fitting \rem{abilities} power of \rem{multi-layer neural} deep networks, \rem{insufficient} limited attention has been \rem{given} paid to\rem{ effectively} integrating high-level tasks\rem{ with low-level vision}. Cross-task learning \rem{enables consolidated models to address challenges arising from limited training data and the absence of ground truth} offers a way to alleviate data scarcity and the lack of ground truth \cite{Zhang_Xu_Xiao_Guo_Ma_2020}. Liu et al. \cite{Liu_Liu_Wu_Ma_Fan_Liu_2023}\rem{ developed a DL model for a dual-task hierarchy, enhancing the mutual benefits of semantic perception and image-fusion tasks.} designed a dual-task hierarchy to jointly enhance semantic perception and image fusion. To emphasize both target-related information and pixel-level details in an image, Sun et al. \cite{sun2022detfusion} leveraged insights from object detection networks to direct MMIF. To explore the potential of image-level fusion in semantically driven methods, Tang et al. \cite{Tang_Zhang_Xu_Ma} introduced a scene-authenticity-constrained image-fusion network with incremental semantic integration. Additionally, Xu et al. \cite{Xu_Ma_Yuan_Le_Liu} leveraged image-fusion responses to enhance registration accuracy. Wang et al. \cite{Wang_Liu_Fan_Liu_2022} proposed a paradigm for handling misaligned infrared and visible light images using intermodal generation and registration. To overcome \rem{the }limitations of two-stage unregistered IVIF, MulFS-CAP \cite{12} introduced a shared shallow feature encoder with a learnable modality dictionary,\rem{ for consistent feature learning, followed by feature reorganization and fusion via a cross-modality alignment mechanism. } enabling consistent feature learning and cross-modality alignment.

\subsection{Benchmarks}

Existing MMIF datasets often \rem{focus on} target specific scenarios. For example, the TNO dataset \cite{ toet2017tno} covers 261 pairs of multi-band images from military, rural, and urban scenes. The RoadScene dataset \cite{Xu_Ma_Le_Jiang_Guo_2020} is dedicated to road scenes, containing 221 pairs of registered images. LLVIP \cite{Jia_Zhu_Li_Tang_Zhou_2021} predominantly features nighttime urban road scenes with 16836 image pairs. The MSRS dataset \cite{Tang_Yuan_Zhang_Jiang_Ma} specializes in multispectral road scenes and includes 1444 pairs of registered images. The M\textsuperscript{3}FD dataset \cite{Liu_Fan_Huang_Wu_Liu_Zhong_Luo} encompasses various lighting conditions and challenging scenarios, totaling 4200 pairs of registered images. Despite their unique characteristics, these datasets are constrained by specific shooting perspectives and locations, posing challenging in fully capturing the complexity and diversity of MMIF in the real world.

Table \ref{table1} summarizes \rem{a comprehensive overview of publicly available Multi-modal datasets} existing public multimodal datasets and the proposed UMS dataset\rem{. This summary includes key aspects: datasets counts, average resolution, shooting perspectives, lighting conditions, objectives, and labels presence.} in terms of counts, resolution, perspectives, lighting, objectives, and label availability.

\section{The proposed TSJNet}

\subsection{Problem formulation}

Unlike prior image-fusion methods that focus solely on visual effects, our approach aims to generate information-rich images that satisfy the dual perception of segmentation and detection requirements. To achieve this, we unify the three tasks into cohesive objective. For illustrative purposes, consider the case of IVIF, where the infrared, visible, and fused images possess dimensions of \( p \times q \). We denote these images as individual vectors \( \mathbf{I} \), \( \mathbf{V} \), \( \mathbf{F} \in \mathbb{R}^{p \times q \times l} \), respectively, with \( l\) representing the number of channels. The optimization framework is defined as follows:

\begin{equation}
\min_{\omega_F, \omega_d, \omega_s} f_F (\mathbf{F}, \Phi(\mathbf{I},\mathbf{V}, \omega_F))+ f_d (\mathbf{d}, \Psi(\mathbf{I}, \mathbf{V},\mathbf{F}, \omega_d)) +
f_s (\mathbf{s}, \varphi(\mathbf{F}, \omega_s))
\end{equation}
where \( f(\cdot) \) denotes the fidelity term. \( \mathbf{F} \), \( \mathbf{d} \), and \( \mathbf{s} \) represent the fusion, detection, and segmentation results, respectively. These are produced by the fusion network \( \Phi \), detection network \( \Psi \) and segmentation network \( \varphi \), with adjustable parameters denoted as \(\omega_F, \omega_d, \omega_s\).
\begin{figure}[h]
  \centering
   \includegraphics[width=1.0\linewidth]{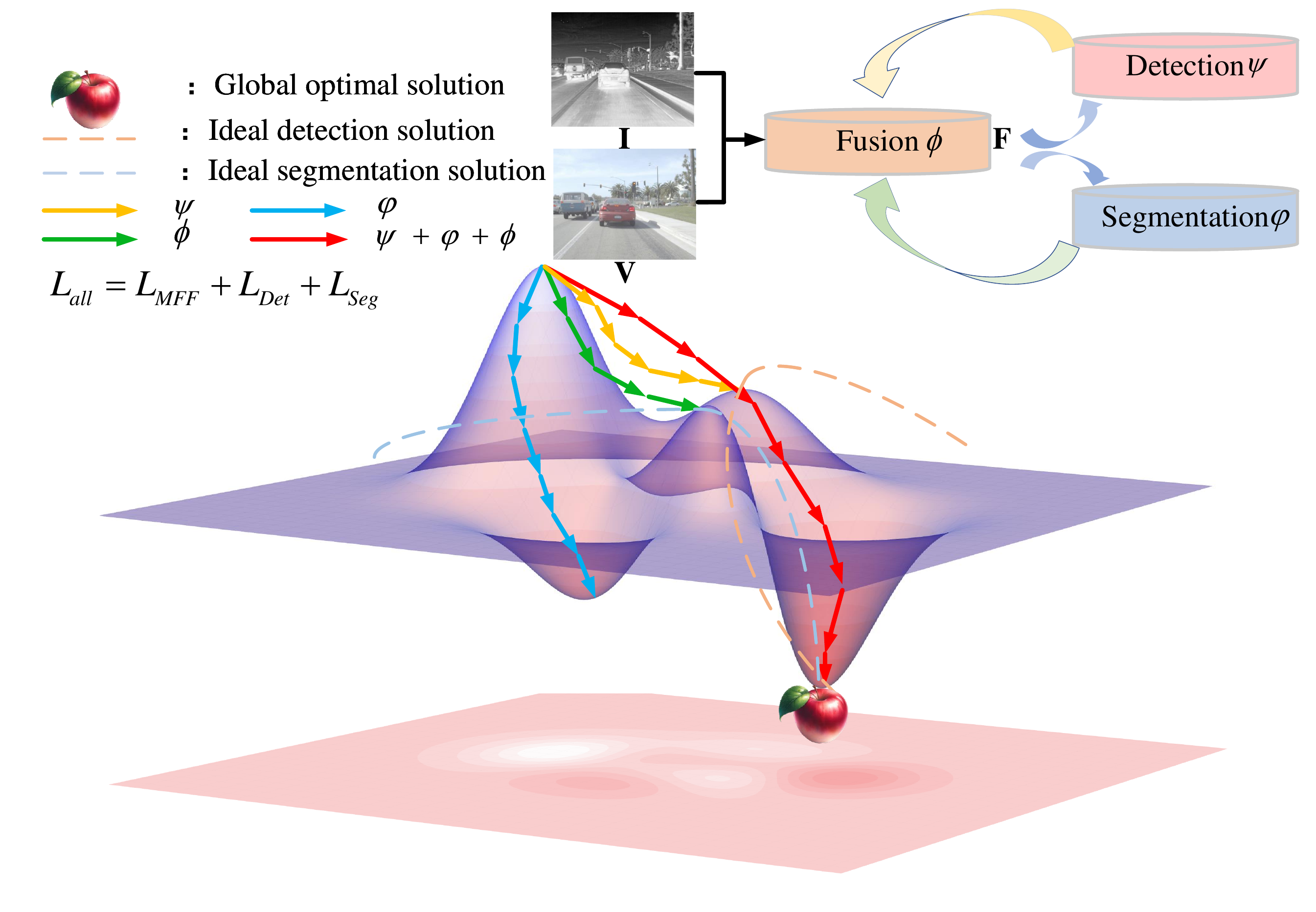}
   \caption{Fusion gradient optimization process. The joint loss design of fusion, detection, and segmentation optimizes the balance among the fusion, detection, and segmentation solutions to achieve the optimal global solution.}
   \label{fig2}
\end{figure}


Fig.~\ref{fig2} illustrates the gradient optimization process for fusion, detection, and segmentation, where the joint loss function balances these components to achieve the global optimal solution. Specifically, we adopt  DeepLabV3+ with ResNet101 \cite{Chen_Papandreou_Schroff_Adam_2017} and Faster R-CNN \cite{Ren_He_Girshick_Sun_2017} as the baseline models for detection and segmentation, providing both target and semantic information.

\begin{figure}[h]
  \centering
   \includegraphics[width=1.0\linewidth]{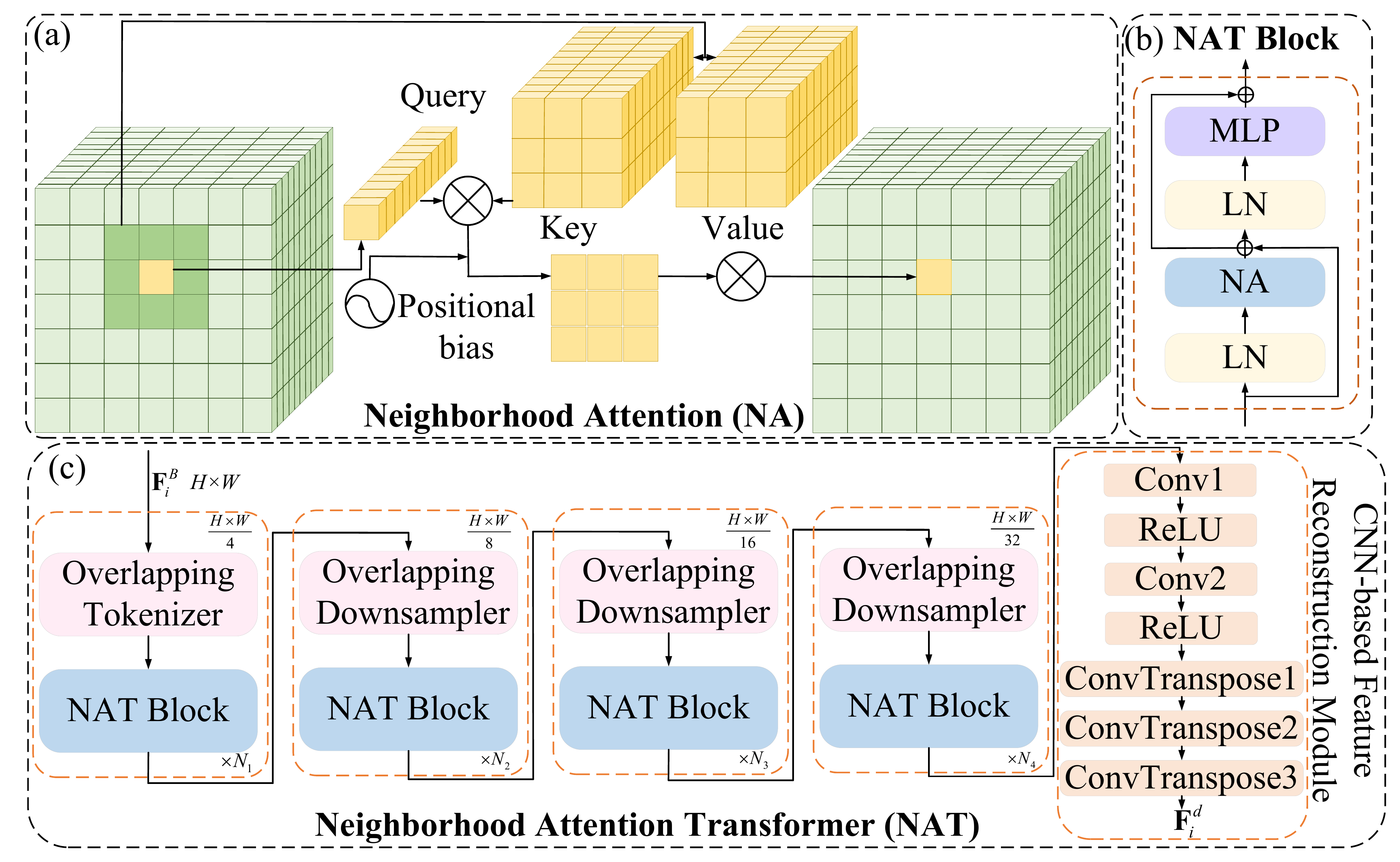}
   \caption{The structure of neighborhood attention transformer.}
   \label{fig_nat}
   \vspace{-10pt}
\end{figure}


\begin{figure*}[htbp]
\centering	
\includegraphics[width=1.0\linewidth]{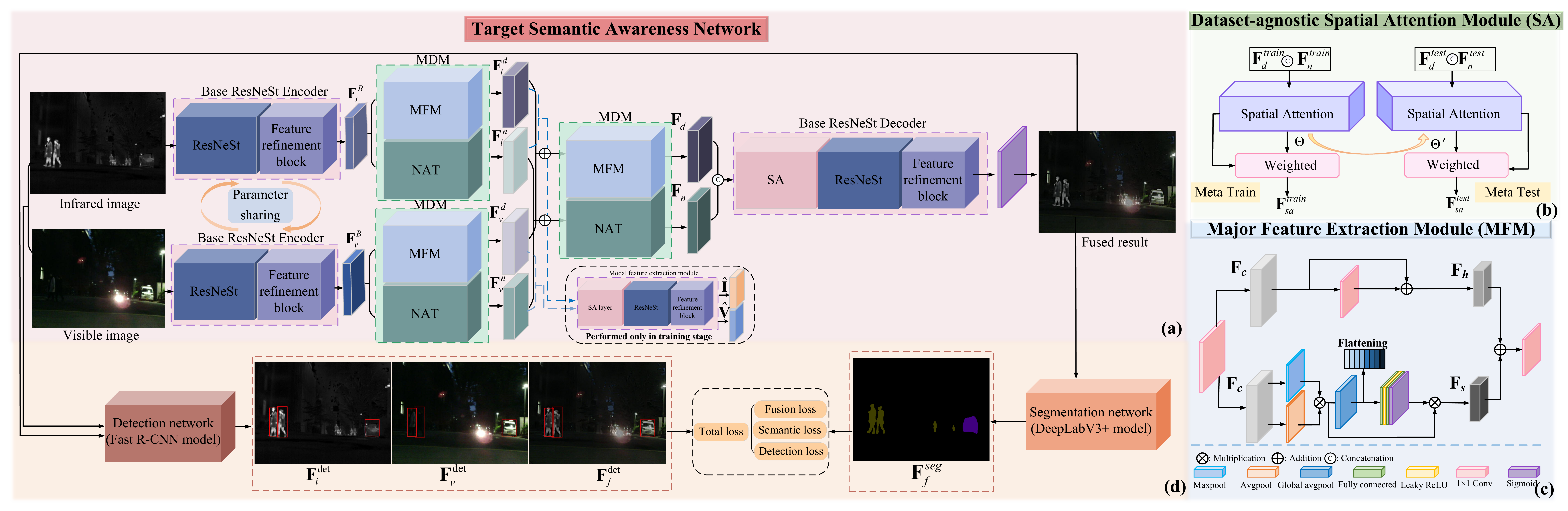}
\caption{Framework of the proposed TSJNet with dual drivers of segmentation and detection. Our model  comprises a base ResNeSt encoder, a dual-branch fusion layer, and a base ResNeSt decoder. The MDM module integrates two parallel branches: NAT, which captures multi-scale contextual dependencies through localized attention and hierarchical design, and MFM, which focuses on extracting salient structures via residual and saliency-enhanced representations.}
\vspace{-10pt} 
\label{Frame}
\end{figure*}

\subsection{Details of TSJNet}
Fig.~\ref{Frame} illustrates the detailed TSJNet model. We introduce symbols to enhance formulation clarity. The paired input images—specifically the infrared and visible images—are denoted as \( \bm{I} \in \mathbb{R}^{p \times q} \) and \( \bm{V} \in \mathbb{R}^{p \times q \times 3} \), respectively.
\subsubsection{\textbf{Base ResNeSt Encoder (BRE)}}
The encoder is comprised two components: a ResNeSt \cite{zhang2022resnest} block and a convolutional layer-based feature refinement block. The \textit{BRE} extracts hierarchical features \( \bm{F^B_I}, \bm{F^B_V} \) from \( \bm{I}, \bm{V} \), that is,
\begin{equation}
    \bm{F^B_i} = BRE(\bm{I}), \quad \bm{F^B_v} = BRE(\bm{V})
\end{equation}

The split-attention of ResNeSt combines channel attention mechanisms with a multi-path network architecture, effectively capturing cross-channel feature correlations. Thus, we integrate ResNeSt into the encoder to learn rich and diverse feature representations of multi-modal images, further enhancing image content understanding.
\subsubsection{\textbf{MDM based fusion layer}} 
To extract and integrate multi-scale shared semantic features and salient modality-specific features from multi-modal images, we propose the MDM. This dual-branch feature extraction unit consists of a Neighborhood Attention Transformer (NAT) \cite{Hassani_Walton_Li_Li_Shi_2022} branch and a Main Feature Module (MFM). As illustrated in Fig.~\ref{fig_nat}, the NAT branch employs a hierarchical structure by stacking NAT blocks to model multi-scale contextual information, effectively capturing both local and global dependencies. In our implementation, the NAT is configured with a neighborhood size of 7×7 following the default design, balancing receptive field coverage and computational efficiency. Additionally, an overlapping tokenizer is introduced to enhance spatial coherence and local continuity of feature embeddings by incorporating overlapping receptive fields during tokenization.

The MFM branch incorporates pooling operations, global feature representations, and a saliency enhancement mechanism. Additionally, the introduction of a residual subnetwork preserves modality-specific attributes, preventing the loss of significant information. As shown in Fig.~\ref{Frame} (c), \( \bm{F_c} \) is generated through a convolutional layer and is processed by residual and salient feature extraction branches. 
The salient feature extraction branch performs average pooling and max pooling for contour and texture information extraction from the background. Dual pooling is followed by global average pooling to extend further image feature representation. Subsequently, the weights for different channels are computed using two fully connected layers and a sigmoid layer, enhancing significant feature descriptions. Summing this branch with a residual branch extracts salient and background information from infrared and visible images. The expressions are as follows:
\begin{equation}
    \{\bm{F^{d}_{i}}, \bm{F^{n}_{i}}\} = MDM(\bm{F^{B}_{i}}), \quad \{\bm{F^{d}_{v}}, \bm{F^{n}_{v}}\} = MDM(\bm{F^{B}_{v}})
\end{equation}
where \( \bm{F^{d}_{i}} \) and \( \bm{F^{n}_{i}} \) represent the salient features of the background and target in \( \bm{I} \). \( \bm{F^{d}_{v}} \) and \( \bm{F^{n}_{v}} \) are for \( \bm{V} \).

By aggregating salient background and target features from \( \bm{I} \) and  \( \bm{V} \), and subsequently inputting them into the MFM and NAT for global-local, shared-special feature extraction, we acquire the pre-fused background and target features, denoted as \( \bm{F_d} \) and \( \bm{F_n} \).
\begin{equation}
    \bm{F_d} = MFM(\bm{F^{d}_{i}} + \bm{F^{d}_{v}}), \quad \bm{F_n} = NAT(\bm{F^{n}_{i}} + \bm{F^{n}_{v}})
\end{equation}
\subsubsection{\textbf{Base ResNeSt Decoder (BRD)}} The design of BRD aligns with the structure of BRE while enhancing the overall stability and robustness of the model. Unlike BRE, BRD incorporates the Dataset-Agnostic Spatial Attention Layer (SA) \cite{Chang_Tong_Du_Hospedales_Song_Ma}.The SA module effectively allocates cross-modal spatial attention weights, enabling the localization of crucial features in the source images across diverse datasets. This attention mechanism facilitates efficient information interaction between multi-modal images, ensuring that the most relevant regions, such as target objects or pathological features, are emphasized for accurate fusion. By highlighting significant areas while maintaining a balanced attention distribution, SA enhances the complementarity and discriminative power of the fused features, reducing redundancy in the model’s information processing. Furthermore, the SA module helps preserve both local details and structural attributes in the fused image, significantly improving the model's generalization ability and performance in downstream tasks, such as object detection and semantic segmentation. During training, SA allows for effective feature interaction and ensures robust performance across various datasets. The pre-fused complementary information extracted by the fusion layer is subsequently used as input to \( BRD(\cdot) \) or image reconstruction, resulting in a high-quality fused output \( \bm{F} \) , which is expressed as follows:

\begin{equation}
    \bm{F}, \bm{att} = BRD(\bm{F_d} + \bm{F_n})
\end{equation}
where \textbf{\textit{att}} is the attention weight matrix generated in the decoder training phase.
\subsection{Loss function}
TSJNet not only enhances the fusion performance but also improves the accuracy of fused images in higher-level tasks by embeding object position and pixel category information into the fusion feature extraction process. Specifically, we combines the detection \( \mathcal{L}_{\text{Det}} \) and segmentation losses \( \mathcal{L}_{\text{Seg}} \) with multifaceted fusion loss \( \mathcal{L}_{\text{MFF}} \). The detection loss is introduced following the formulation in \cite{Ren_He_Girshick_Sun_2017}, while the segmentation loss employs a cross-entropy function. The precise formulation of the fusion loss can be presented as:
\begin{equation}
    \mathcal{L}_{\text{all}} = \mathcal{L}_{\text{MFF}} + \mathcal{L}_{\text{Det}} + \mathcal{L}_{\text{Seg}}
\end{equation}
where \( \mathcal{L}_{\text{all}} \) denotes the total loss.
\subsubsection{MultiFacet Fusion Loss}
The quality of fused images directly impacts the accuracy of advanced tasks. In object detection, precise localization and recognition rely on the complete preservation of object edges and feature information. For semantic segmentation, pixel-level classification requires rich visual details and background information. Accordingly, MFFLoss optimizes fused images to further enhance the overall performance of \rem{object }detection and \rem{semantic }segmentation tasks.

\( \mathcal{L}_{\text{MFF}} \) prioritize preserving target texture, edge information, and background continuity via the loss of the structural similarity metric, \( \mathcal{L}_{\text{ssim}} \), which is critical for pixel-level classification in semantic segmentation, as expressed in Eq.\ref{Lssim}.
\begin{equation}
     \mathcal{L}_{\text{ssim}} = \frac{1}{2} \left(1 - \text{ssim}(\bm{F}, \bm{I})\right) + \frac{1}{2} \left(1 - \text{ssim}(\bm{F}, \bm{V})\right)\label{Lssim} 
\end{equation}
where \(\text{ssim}(\cdot)\) calculates the structural similarity.

To enhance the generalization of the fusion model and improve its capability in optimizing multi-modal image features, we introduce the diversity loss \( \mathcal{L}_{\text{div}} \) \cite{Chang_Tong_Du_Hospedales_Song_Ma}. Specifically, \( \mathcal{L}_{\text{div}} \) imposes constraints on the attention weight matrix of SA, ensuring the fusion model focuses more effectively on the critical regions within the multi-modal image, as expressed in Eq.\ref{div}.
\begin{equation}
\mathcal{L}_{\text{div}} = -\frac{1}{m} \sum_{i=1}^{m} \left(1 - \max_{j} \bm{att}_{i,j}\right) + \frac{1}{mn} \sum_{i=1}^{m} \sum_{j=1}^{n} \bm{att}_{i,j}\label{div}
\end{equation}
where and $m$ and $n$ are the number of rows and columns, respectively.

In summary, the objective of our model can be succinctly summarized as follows:
\begin{equation}
\mathcal{L}_{\text{MFF}} = \alpha_1 \mathcal{L}_{\text{div}} +  \mathcal{L}_{\text{ssim}} + \alpha_2 \mathcal{L}'_{\text{mse}} + \alpha_3 \mathcal{L}''_{\text{mse}}
\label{MFF}
\end{equation}
where \( \mathcal{L}'_{\text{mse}} = \| \bm{I} - \bm{\hat{I}} \|_2^2 \) and \( \mathcal{L}''_{\text{mse}} = \| \bm{V} - \bm{\hat{V}} \|_2^2 \), $\hat{\bm{I}}$ and $\hat{\bm{V}}$ are outputs of BRD. \( \alpha_1 \), \( \alpha_2 \), and \( \alpha_3 \) are the tunning parameters.
\subsubsection{Detection Loss}

The infrared, visible, and fused results independently feed into the detection network for detection loss computation. We represent the detection loss for the \(\bm{I}\), \(\bm{V}\), and \(\bm{F}\) as \( \mathcal{L}^I_{\text{det}} \), \( \mathcal{L}^V_{\text{det}} \), and \( \mathcal{L}^F_{\text{det}} \), respectively. The detection loss \( \mathcal{L}_{\text{Det}} \) can be denoted as follows,

\begin{equation}
\mathcal{L}_{\text{Det}} = \mathcal{L}^I_{\text{det}} + \mathcal{L}^V_{\text{det}} + \mathcal{L}^F_{\text{det}}
\end{equation}
To enhance computational efficiency of our model, we eliminated the classification loss from the original object detection loss \cite{Ren_He_Girshick_Sun_2017}, retaining solely the regression loss \cite{sun2022detfusion}. Furthermore, the detection loss function works only when the fusion network needs to be modified. Consider a fusion-detection network, where example,
\begin{equation}
\mathcal{L}^F_{\text{det}}(u, t^u, v) = [u \geq 1]\mathcal{L}^F_{\text{det}}(t^u, v)
\end{equation}
where \( u \) is the ground truth class, \( t^u \) is the the predicted outcome of the bounding box, and \( v \) is the regression goal of the ground truth bounding box. The Iverson bracket indicator function \([u \geq 1]\)
is 1 when \( u \geq 1 \) is satisfied, and 0 otherwise. We label the background class as \( u = 0 \).
\begin{table*}[h!]
\caption{Objective comparisons of different methods on MSRS, M\textsuperscript{3}FD, RoadScene, LLVIP, TNO, and the proposed UMS datasets.The top three values are marked in \textcolor{top1}{red}, \textcolor{top2}{green}, and \textcolor{top3}{blue}.}
\label{tab:comparison}
\resizebox{\textwidth}{!}{%
\begin{tabular}{lcccccc|lcccccc}
\hline
\multicolumn{7}{c|}{MSRS \cite{Tang_Yuan_Zhang_Jiang_Ma} dataset} &
  \multicolumn{7}{c}{M\textsuperscript{3}FD \cite{Liu_Fan_Huang_Wu_Liu_Zhong_Luo} dataset} \\
Method &
  {SSIM$\uparrow$} &
  {PSNR$\uparrow$} &
  {MSE$\downarrow$} &
  {VIF$\uparrow$} &
  {CC$\uparrow$} &
  {CV$\downarrow$} &
Method &
  {SSIM$\uparrow$} &
  {PSNR$\uparrow$} &
  {MSE$\downarrow$} &
  {VIF$\uparrow$} &
  {CC$\uparrow$} &
  {CV$\downarrow$} \\ \hline
ReC \cite{Huang_Liu_Fan_Liu_Zhong_Luo}&
  0.30 &
  16.38 &
  2005.55 &
  0.31 &
  0.56 &
  \textcolor{top2}{319.06} &
  ReC \cite{Huang_Liu_Fan_Liu_Zhong_Luo} &
  0.67 &
  \textcolor{top3}{14.11} &
  \textcolor{top3}{2989.60} &
  0.33 &
  0.51 &
  {\color{top1}496.54} \\
CDD \cite{Zhao_Bai_Zhang_Zhang_Xu_Lin_Timofte_Gool_2022} &
  \textcolor{top2}{0.68} &
  16.23 &
  2437.54 &
  \textcolor{top2}{0.44} &
  \textcolor{top3}{0.60} &
  \textcolor{top1}{230.35} &
  CDD \cite{Zhao_Bai_Zhang_Zhang_Xu_Lin_Timofte_Gool_2022} &
  \textcolor{top2}{0.69} &
  13.04 &
  4035.47 &
  {\color{top3}0.37} &
  0.52 &
  \textcolor{top2}{501.84} \\
LRR \cite{Li_Xu_Wu_Lu_Kittler_2023} &
  {0.59} &
  16.61 &
  2039.84 &
  0.33 &
  0.51 &
  613.42 &
  LRR \cite{Li_Xu_Wu_Lu_Kittler_2023} &
  {\color{top1}0.70} &
  \textcolor{top2}{14.48} &
  \textcolor{top2}{2722.86} &
  0.36 &
  {\color{top2}0.53} &
  678.53 \\
TarD \cite{Liu_Fan_Huang_Wu_Liu_Zhong_Luo} &
  0.46 &
  \textcolor{top3}{16.99} &
  \textcolor{top3}{1941.00} &
  {0.34} &
  0.46 &
  2342.74 &
  TarD \cite{Liu_Fan_Huang_Wu_Liu_Zhong_Luo} &
  0.68 &
  13.74 &
  3385.62 &
  \textcolor{top2}{0.39} &
  0.44 &
  1271.50 \\
IGN \cite{Li_Chen_Liu_Ma_2023} &
  0.56 &
  \textcolor{top2}{17.22} &
  \textcolor{top2}{1811.24} &
  0.32 &
  \textcolor{top1}{0.65} &
  903.80 &
  IGN \cite{Li_Chen_Liu_Ma_2023} &
  0.58 &
  12.11 &
  4649.93 &
  0.24 &
  0.52 &
  1020.07 \\
Sup \cite{Tang_Deng_Ma_Huang_Ma_2022} &
  0.50 &
  14.05 &
  4717.51 &
  0.24 &
  0.27 &
  1667.85 &
  Sup \cite{Tang_Deng_Ma_Huang_Ma_2022} &
  {\color{top2}0.69} &
  12.98 &
  4019.42 &
  {\color{top1}0.40} &
  0.47 &
  598.91 \\
DIV \cite{Tang_Xiang_Zhang_Gong_Ma} &
  0.34 &
  8.51 &
  9613.72 &
  0.18 &
  0.57 &
  1364.19 &
  DIV \cite{Tang_Xiang_Zhang_Gong_Ma} &
  0.59 &
  12.30 &
  3957.61 &
  0.23 &
  {\color{top1}0.54} &
  871.24 \\
EMM \cite{06}  &
  \textcolor{top3}{0.66} &
  16.24 &
  2441.76 &
  0.39 &
  0.59 &
  354.38 &
  EMM \cite{06} &
  \textcolor{top2}{0.69} &
  13.12 &
  3994.92 &
  0.33 &
  0.50 &
  525.45 \\
SHI \cite{07} &
  \textcolor{top3}{0.66} &
  15.93 &
  2560.61 &
  \textcolor{top3}{0.40} &
  0.59 &
  350.71 &
  SHI \cite{07} &
  0.68 &
  12.74 &
  4155.72 &
  0.35 &
  0.45 &
  546.44 \\
\textbf{TSJNet} &
  \textcolor{top1}{0.69} &
  \textcolor{top1}{18.13} &
  \textcolor{top1}{1502.60} &
  \textcolor{top1}{0.45} &
  \textcolor{top2}{0.62} &
  \textcolor{top3}{344.20} &
  \textbf{TSJNet} &
  {\color{top1}0.70} &
  {\color{top1}14.63} &
  {\color{top1}2714.86} &
  0.33 &
  {\color{top2}0.53} &
  {\color{top3}524.41} \\ \hline
\multicolumn{7}{c|}{RoadScene \cite{Xu_Ma_Le_Jiang_Guo_2020} dataset} &
  \multicolumn{7}{c}{LLVIP \cite{Jia_Zhu_Li_Tang_Zhou_2021} dataset} \\
 Method &
  {SSIM$\uparrow$} &
  {PSNR$\uparrow$} &
  {MSE$\downarrow$} &
  {VIF$\uparrow$} &
  {CC$\uparrow$} &
  {CV$\downarrow$} &
   Method &
  {SSIM$\uparrow$} &
  {PSNR$\uparrow$} &
  {MSE$\downarrow$} &
  {VIF$\uparrow$} &
  {CC$\uparrow$} &
  {CV$\downarrow$} \\ \hline
ReC \cite{Huang_Liu_Fan_Liu_Zhong_Luo} &
  {\color{top2}0.72} &
  {\color{top1}15.40} &
  {\color{top1}2404.92} &
  \textcolor{top3}{0.32} &
  0.62 &
  {\color{top3}460.00} &
  ReC \cite{Huang_Liu_Fan_Liu_Zhong_Luo} &
  0.42 &
  14.64 &
  2281.86 &
  0.30 &
  {\color{top1}0.69} &
  {\color{top1}302.05} \\
CDD \cite{Zhao_Bai_Zhang_Zhang_Xu_Lin_Timofte_Gool_2022} &
  0.67 &
  14.03 &
  3077.02 &
  0.25 &
  {\color{top2}0.63} &
  \textcolor{top2}{426.30} &
  CDD \cite{Zhao_Bai_Zhang_Zhang_Xu_Lin_Timofte_Gool_2022} &
  {\color{top2}0.64} &
  14.58 &
  2315.02 &
  {\color{top1}0.41} &
  \textcolor{top2}{0.68} &
  \textcolor{top2}{332.29} \\
LRR \cite{Li_Xu_Wu_Lu_Kittler_2023} &
  0.58 &
  11.82 &
  4401.71 &
  0.24 &
  \textcolor{top3}{0.62} &
  629.70 &
  LRR \cite{Li_Xu_Wu_Lu_Kittler_2023} &
  0.59 &
  {\color{top2}15.93} &
  {\color{top1}1688.62} &
  \textcolor{top3}{0.39} &
  \textcolor{top2}{0.68} &
  580.90 \\
TarD \cite{Liu_Fan_Huang_Wu_Liu_Zhong_Luo} &
  0.69 &
  {\color{top2}14.82} &
  \textcolor{top3}{2603.71} &
  0.31 &
  0.58 &
  1255.83 &
  TarD \cite{Liu_Fan_Huang_Wu_Liu_Zhong_Luo} &
  0.56 &
  14.33 &
  2508.97 &
  {\color{top1}0.41} &
  0.65 &
  1095.24 \\
IGN \cite{Li_Chen_Liu_Ma_2023} &
  0.52 &
  10.31 &
  6693.45 &
  0.28 &
  0.60 &
  1033.15 &
  IGN \cite{Li_Chen_Liu_Ma_2023} &
  0.55 &
  \textcolor{top3}{14.97} &
  \textcolor{top3}{2137.50} &
  0.24 &
  \textcolor{top2}{0.68} &
  638.59 \\
Sup \cite{Tang_Deng_Ma_Huang_Ma_2022} &
  {\color{top1}0.74} &
  14.51 &
  2928.35 &
  {\color{top1}0.34} &
  0.60 &
  {\color{top1}420.00} &
  Sup \cite{Tang_Deng_Ma_Huang_Ma_2022} &
  {\color{top1}0.64} &
  14.62 &
  2306.66 &
  0.38 &
  \textcolor{top2}{0.68} &
  {\color{top3}360.89} \\
DIV \cite{Tang_Xiang_Zhang_Gong_Ma} &
  0.61 &
  13.86 &
  3033.92 &
  0.19 &
  0.62 &
  907.78 &
  DIV \cite{Tang_Xiang_Zhang_Gong_Ma} &
  0.46 &
  10.44 &
  6362.27 &
  0.25 &
  \textcolor{top2}{0.68} &
  645.38 \\
EMM \cite{06} &
  0.66 &
  14.04 &
  3037.58 &
  0.25 &
  \textcolor{top3}{0.62} &
  416.33 &
  EMM \cite{06} &
  0.64 &
  14.59 &
  2320.60 &
  0.35 &
  0.68 &
  310.12\\
SHI \cite{07} &
  0.66 &
  13.78 &
  3364.21 &
  0.27 &
  0.57 &
  425.21 &
  SHI \cite{07} &
   0.62&
   13.94&
   2697.28&
   0.35&
   0.66&
   374.81\\
\textbf{TSJNet} &
  {\color{top3}0.70} &
  {\color{top3}14.52} &
  {\color{top2}2544.58} &
  {\color{top2}0.33} &
  {\color{top1}0.64} &
  516.82 &
  \textbf{TSJNet} &
  {\color{top1}0.64} &
  {\color{top1}15.97} &
  {\color{top2}1697.69} &
  0.37 &
  0.67 &
  457.46 \\ \hline
\multicolumn{7}{c|}{TNO \cite{toet2017tno} dataset} &
  \multicolumn{7}{c}{UMS dataset} \\
 Method &
  {SSIM$\uparrow$} &
  {PSNR$\uparrow$} &
  {MSE$\downarrow$} &
  {VIF$\uparrow$} &
  {CC$\uparrow$} &
  {CV$\downarrow$} &
   Method &
  {SSIM$\uparrow$} &
  {PSNR$\uparrow$} &
  {MSE$\downarrow$} &
  {VIF$\uparrow$} &
  {CC$\uparrow$} &
  {CV$\downarrow$} \\ \hline
ReC \cite{Huang_Liu_Fan_Liu_Zhong_Luo} &
  {\color{top2}0.73} &
  {\color{top1}17.67} &
  {\color{top1}1598.49} &
  0.36 &
  0.60 &
  {\color{top3}337.37} &
  ReC \cite{Huang_Liu_Fan_Liu_Zhong_Luo} &
  0.61 &
  {\color{top1}13.95} &
  {\color{top1}2730.45} &
  0.31 &
  {\color{top1}0.58} &
  545.29 \\
CDD \cite{Zhao_Bai_Zhang_Zhang_Xu_Lin_Timofte_Gool_2022} &
  0.67 &
  14.88 &
  2631.73 &
  0.30 &
  {\color{top2}0.61} &
  \textcolor{top1}{276.78} &
  CDD \cite{Zhao_Bai_Zhang_Zhang_Xu_Lin_Timofte_Gool_2022} &
  {\color{top2}0.65} &
  12.59 &
  3753.37 &
  {\color{top2}0.34} &
  \textcolor{top1}{0.58} &
  \textcolor{top2}{397.27} \\
LRR \cite{Li_Xu_Wu_Lu_Kittler_2023} &
  0.64 &
  14.42 &
  2895.15 &
  0.27 &
  0.57 &
  501.70 &
  LRR \cite{Li_Xu_Wu_Lu_Kittler_2023} &
  0.60 &
  {\color{top2}13.84} &
  {\color{top2}2912.62} &
  0.23 &
  \textcolor{top2}{0.57} &
  1051.11 \\
TarD \cite{Liu_Fan_Huang_Wu_Liu_Zhong_Luo} &
  {\color{top2}0.73} &
  {\color{top2}17.32} &
  \textcolor{top2}{1861.51} &
  {\color{top2}0.37} &
  0.57 &
  594.18 &
  TarD \cite{Liu_Fan_Huang_Wu_Liu_Zhong_Luo} &
  0.40 &
  12.09 &
  4128.53 &
  0.14 &
  0.47 &
  666.21 \\
IGN \cite{Li_Chen_Liu_Ma_2023} &
  0.44 &
  11.45 &
  5656.93 &
  0.15 &
  0.30 &
  1146.83 &
  IGN \cite{Li_Chen_Liu_Ma_2023} &
  0.56 &
  13.10 &
  3297.93 &
  0.24 &
  0.55 &
  {\color{top1}371.82} \\
Sup \cite{Tang_Deng_Ma_Huang_Ma_2022} &
  {\color{top1}0.75} &
  {\color{top3}16.46} &
  {\color{top3}2143.23} &
  {\color{top1}0.39} &
  0.59 &
  {\color{top2}287.04} &
  Sup \cite{Tang_Deng_Ma_Huang_Ma_2022} &
  {\color{top2}0.65} &
  12.80 &
  3691.07 &
  {\color{top1}0.35} &
  0.52 &
  436.49 \\
DIV \cite{Tang_Xiang_Zhang_Gong_Ma} &
  0.58 &
  13.05 &
  3401.38 &
  0.22 &
  {\color{top2}0.61} &
  596.28 &
  DIV \cite{Tang_Xiang_Zhang_Gong_Ma} &
  0.59 &
  11.92 &
  4303.60 &
  0.25 &
  \textcolor{top2}{0.57} &
  657.34 \\
EMM \cite{06} &
  0.65 &
  15.11 &
  2445.93 &
  0.28 &
  0.59 &
  271.68 &
  EMM \cite{06} &
  0.62 &
  13.57 &
  3002.12 &
  0.30 &
  0.57 &
  491.85 \\
SHI \cite{07} &
  0.70 &
  16.32 &
  2350.22 &
  0.33 &
  0.56 &
  246.16 &
  SHI \cite{07} &
  0.60 &
  12.05 &
  4459.79 &
  0.29 &
  0.49 &
  432.90 \\

\textbf{TSJNet} &
  0.72 &
  14.93 &
  2595.32 &
  {\color{top2}0.37} &
  {\color{top1}0.63} &
  {\color{top3}311.50} &
  \textbf{TSJNet} &
  {\color{top1}0.66} &
  {\color{top3}13.75} &
  {\color{top3}2992.80} &
  {\color{top2} 0.34} &
  0.55 &
  {\color{top3}415.43} \\ \hline
\end{tabular}%
}
\end{table*}
\begin{figure*}[h!]
\centering	
\includegraphics[width=1.0\linewidth]{04subjective_assessment3-16.pdf}
\caption{Subjective comparisons of different methods on MSRS, M\textsuperscript{3}FD, RoadScene, LLVIP, TNO, and \rem{the proposed} UMS datasets.}
\vspace{-10pt}
\label{04}
\end{figure*}
\begin{figure}[htbp]
\centering	
\includegraphics[width=1\linewidth]{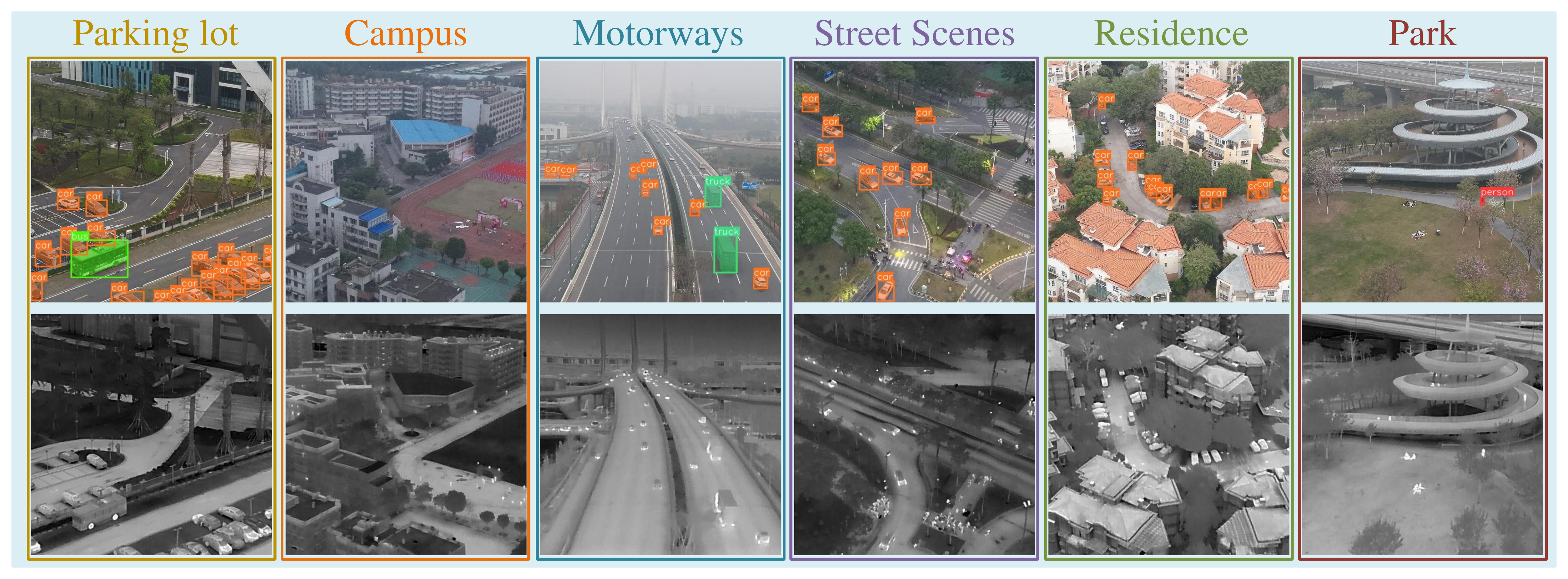}
\caption{\del{Visualization of the proposed UMS dataset.} The proposed UMS dataset is visualized across six distinct scenarios: parking lots, parks, campuses, motorways, city street scenes, and residential areas.} 
\vspace{-10pt}
\label{fig.9}
\end{figure}

\subsubsection{Semantic Loss}
Incorporating DeepLabV3+ \cite{Chen_Papandreou_Schroff_Adam_2017} into our model significantly enhances the semantic information of the fusion results. The semantic loss can be expressed as:
\begin{equation}
\mathcal{L}_{\text{seg}} = -\frac{1}{PQ} \sum_{p=1}^{P} \sum_{q=1}^{Q} \log\left(\frac{\exp(\hat{y}_{c_{p,q},p,q})}{\sum_{j=1}^{C} \exp(\hat{y}_{j,p,q})}\right)
\end{equation}
where \( c_{p,q} \) represents the true class index at position \((p, q)\). \( \hat{y}_{j,p,q} \) signifies the prediction of the model for class \( j \) at position \((p, q)\) in logits.

\section{UAV multi-scenario Benchmark}
Conventional infrared and visible datasets, captured by ground-based fixed cameras with limited viewpoints, struggle to represent the real-world complexity and diversity. In contrast, the UMS dataset, leveraging aerial photography techniques, encompasses a broad spectrum of scenarios and target types, rendering it highly valuable for research.


Fig.~\ref{fig.8} illustrates the UMS dataset captured using a DJI Matrice M30T UAV. This UAV is proficient in stable night and windy conditions for flight and photography. Its infrared camera operates within a spectral range of 8-14$\mu m$. The optical centers of the infrared and visible cameras are 3 cm apart. The captured visible and infrared images have resolutions of 4000×3000 and 1280×1024, respectively. Additionally, the visible camera boasts a 21 mm focal length and a variable aperture (ranging from f/2.8 to f/4.2), enabling precise depth of field and brightness control. In contrast, the infrared sensor has a fixed aperture of f/1.0 and a 9.1 mm focal length, ensuring efficient IR spectral range capture and clear imaging. To construct this dataset, a series of flight paths were meticulously designed to facilitate data collection across diverse real-world scenarios in Foshan, China. All data collection activities were conducted in strict accordance with local regulations. The entire process was collaboratively completed by three researchers, who were respectively responsible for drone operation, simulating human targets during data acquisition, and organizing and integrating the collected data. The dataset construction spanned approximately two months.


To process the UMS dataset, visible images undergo initial cropping to match the size and view of infrared images. Subsequently, both image types align using the alignment algorithm of SuperFusion \cite{Tang_Deng_Ma_Huang_Ma_2022}. Each aligned image pair has a resolution of 640 $\times$ 512 and the dataset includes data captured from three distinct viewpoints: ground, drive, and flight. The UMS dataset comprises 346 image sets across six scenarios, including parking lot, park, campus, motorways, city street scenes, and residence, as depicted in Fig.~\ref{fig.9}. Additionally, we furnish labels for target detection and semantic segmentation, establishing a foundation for investigating image-fusion algorithms’ impact on these tasks.

\section{Experiment}

\subsection{Setup}
\begin{figure*}[htbp]
\centering	
\includegraphics[width=1.0\linewidth]{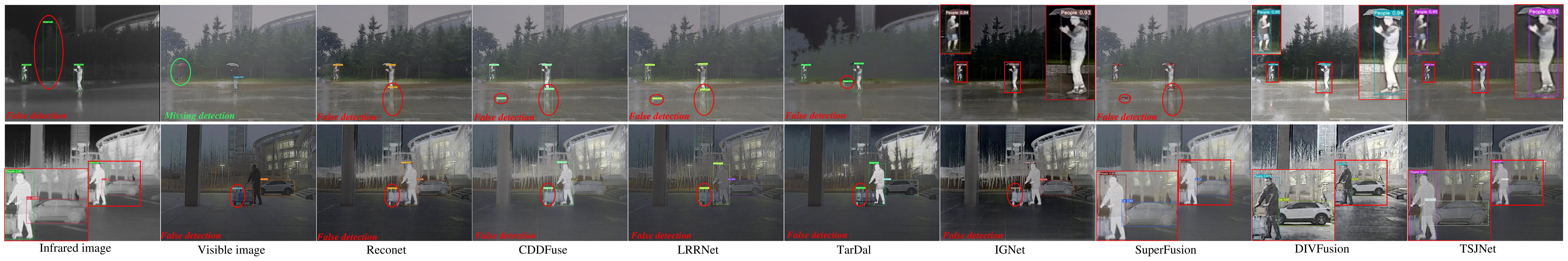}
\caption{Detection results of different methods on M\textsuperscript{3}FD dataset.}
\label{object}
\end{figure*}
\begin{figure*}[htbp]
\centering	
\includegraphics[width=1.0\linewidth]{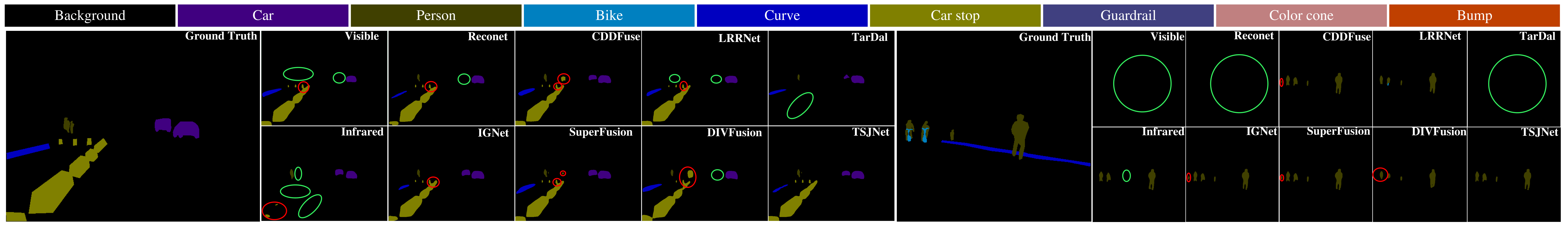}
\caption{Segmentation results \rem{of different methods }on the MSRS dataset, with missing and incorrect \rem{parts are marked with} regions highlighted in} green and red ovals.
\label{segmentation}
\end{figure*}
\begin{table*}[htbp]
\caption{Comparison of TSJNet with seven methods for semantic segmentation on the MSRS and object detection on the M\textsuperscript{3}FD. (The top three values are marked in \textcolor{top1}{red}, \textcolor{top2}{green}, and \textcolor{top3}{blue}.})
\label{tab:seg/det comparisons}
\resizebox{\textwidth}{!}{%
\begin{tabular}{l|cccccccccc|ccccccc}
\hline
\multicolumn{1}{c|}{} &
  \multicolumn{10}{c|}{IoU (Dataset: MSRS \cite{Tang_Yuan_Zhang_Jiang_Ma})} &
  \multicolumn{7}{c}{AP@0.5 (Dataset: M\textsuperscript{3}FD \cite{Liu_Fan_Huang_Wu_Liu_Zhong_Luo})} \\
 Method &
  {Background} &
  {Car} &
  {Person} &
  {Bike} &
  {Curve} &
  {Car Stop} &
  {Guardrail} &
  {Color Tone} &
  {Bump} &
  {mIoU} &
  {Peo} &
  {Car} &
  {Bus} &
  {Mot} &
  {Lam} &
  {Tru} &
  {mAP@.5} \\ \hline
Visible &
  97.92 &
  86.79 &
  39.97 &
  70.50 &
  53.33 &
  71.85 &
  \textcolor{top1}{85.90} &
  \textcolor{top2}{65.44} &
  \textcolor{top3}{79.16} &
  0.7232 &
  0.803 &
  0.912 &
  0.946 &
  0.841 &
  0.667 &
  \textcolor{top3}{0.890} &
  0.843 \\
Infrared &
  96.14 &
  61.90 &
  70.00 &
  24.46 &
  33.64 &
  20.67 &
  0.00 &
  20.98 &
  27.97 &
  0.3953 &
  0.777 &
  0.733 &
  0.597 &
  0.113 &
  0.519 &
  0.611 &
  0.558 \\
ReC \cite{Huang_Liu_Fan_Liu_Zhong_Luo} &
  97.57 &
  83.08 &
  56.20 &
  58.06 &
  37.91 &
  57.34 &
  77.66 &
  55.83 &
  59.68 &
  0.6481 &
  0.803 &
  0.912 &
  0.946 &
  0.841 &
  0.667 &
  0.890 &
  0.843 \\
CDD \cite{Zhao_Bai_Zhang_Zhang_Xu_Lin_Timofte_Gool_2022} &
  97.64 &
  \textcolor{top1}{90.28} &
  72.37 &
  \textcolor{top3}{72.06} &
  \textcolor{top2}{63.32} &
  \textcolor{top3}{73.27} &
  82.05 &
  64.34 &
  \textcolor{top1}{80.57} &
  \textcolor{top2}{0.7732} &
  \textcolor{top3}{0.832} &
  \textcolor{top3}{0.916} &
  \textcolor{top3}{0.953} &
  \textcolor{top2}{0.844} &
  0.705 &
  0.882 &
  \textcolor{top3}{0.855} \\
LRR \cite{Li_Xu_Wu_Lu_Kittler_2023} &
  \textcolor{top3}{98.31} &
  88.85 &
  67.21 &
  69.56 &
  52.12 &
  71.51 &
  81.07 &
  63.93 &
  77.63 &
  0.7446 &
  0.796 &
  \textcolor{top1}{0.923} &
  \textcolor{top2}{0.955} &
  \textcolor{top2}{0.866} &
  \textcolor{top2}{0.721} &
  \textcolor{top1}{0.899} &
  \textcolor{top1}{0.860} \\
TarD \cite{Liu_Fan_Huang_Wu_Liu_Zhong_Luo} &
  97.52 &
  82.12 &
  55.17 &
  63.56 &
  38.83 &
  63.94 &
  58.58 &
  53.15 &
  45.89 &
  0.6208 &
  0.826 &
  0.897 &
  0.918 &
  0.749 &
  0.679 &
  0.874 &
  0.824 \\
IGN \cite{Li_Chen_Liu_Ma_2023} &
  98.46 &
  89.48 &
  \textcolor{top1}{74.01} &
  70.76 &
  57.69 &
  \textcolor{top3}{73.62} &
  83.59 &
  64.24 &
  72.96 &
  0.7609 &
  \textcolor{top1}{0.849} &
  0.894 &
  0.917 &
  0.750 &
  0.650 &
  0.866 &
  0.821 \\
Sup \cite{Tang_Deng_Ma_Huang_Ma_2022} &
  96.43 &
  88.24 &
  \textcolor{top3}{73.21} &
  \textcolor{top1}{72.22} &
  \textcolor{top3}{62.77} &
  73.20 &
  82.25 &
  65.18 &
  \textcolor{top2}{80.15} &
  0.7709 &
  \textcolor{top2}{0.837} &
  0.910 &
  0.932 &
  0.774 &
  0.700 &
  0.858 &
  0.835 \\
DIV \cite{Tang_Xiang_Zhang_Gong_Ma} &
  98.05 &
  87.38 &
  64.63 &
  67.02 &
  50.63 &
  69.43 &
  78.37 &
  61.47 &
  44.38 &
  0.6904 &
  0.726 &
  0.898 &
  0.925 &
  0.792 &
  0.638 &
  0.887 &
  0.811 \\
EMM \cite{07} &
  \textcolor{top1}{98.58} &
  \textcolor{top2}{90.00} &
  72.92 &
  71.10 &
  \textcolor{top1}{64.38} &
  \textcolor{top1}{74.48} &
  \textcolor{top3}{84.11} &
  \textcolor{top1}{65.68} &
  74.32 &
  \textcolor{top3}{0.7729} &
  0.825 &
  0.826 &
  0.820 &
  0.744 &
  \textcolor{top3}{0.720} &
  0.577 &
  0.752 \\
SHI \cite{06} &
  98.52 &
  89.54 &
  72.87 &
  71.55 &
  \textcolor{top2}{63.69} &
  73.26 &
  \textcolor{top2}{84.76} &
  \textcolor{top2}{65.44} &
  74.58 &
  0.7713 &
  0.821 &
  0.817 &
  0.810 &
  0.742 &
  \textcolor{top1}{0.874} &
  0.497 &
  0.760 \\
\textbf{TSJNet} &
  \textcolor{top2}{98.55} &
  \textcolor{top3}{89.97} &
  \textcolor{top2}{73.73} &
  \textcolor{top3}{71.32} &
  \textcolor{top3}{63.59} &
  \textcolor{top2}{73.90} &
  82.71 &
  64.69 &
  77.48 &
  \textcolor{top1}{0.7733} &
  0.818 &
  \textcolor{top2}{0.918} &
  \textcolor{top1}{0.956} &
  \textcolor{top1}{0.868} &
  \textcolor{top3}{0.704} &
  \textcolor{top2}{0.893} &
  \textcolor{top1}{0.860} \\ \hline
\end{tabular}%
}
\end{table*}
\subsubsection{Experimental Detail}
Experiments were conducted on a server equipped with two RTX 3090 GPUs. We trained using the original image size, avoiding chunking and resizing to preserve semantic information and avoid labeling failure. \del{The training involved 40 epochs with a batch size of 2.} The training was carried out for 40 epochs with a batch size of 2, and the total training time was approximately 19.2 hours. The Adam optimization function was employed with an initial learning rate of 0.001. When loss function stalled for three consecutive times, the learning rate was reduced by a factor of 0.1. For object detection and semantic segmentation, we utilized pre-trained Fast R-CNN models with ResNet-50 and FPN \cite{Ren_He_Girshick_Sun_2017}, along with a pre-trained DeepLabV3+ model using ResNet101 \cite{Chen_Papandreou_Schroff_Adam_2017}. The three adjustable parameters \( \alpha_1 \), \( \alpha_2 \), and \( \alpha_3 \) in Eq.\ref{MFF} are set as 0.1, 6, and 1, respectively. \rem{Moreover, a precision \cite{Micikevicius_Narang_Alben_Diamos_Elsen_Garcia_Ginsburg_Houston_Kuchaiev_Venkatesh_etal} strategy was applied during training to reduce memory usage.}
\subsubsection{Datasets, Metrics, and Compared methods}
\textbf{Datasets}:  Our model was trained on the MSRS dataset \cite{Tang_Yuan_Zhang_Jiang_Ma} (1035 pairs) and was subsequently tested on MSRS (361 pairs), M\textsuperscript{3}FD \cite{Liu_Fan_Huang_Wu_Liu_Zhong_Luo} (300 pairs), RoadScene \cite{Xu_Ma_Le_Jiang_Guo_2020} (221 pairs), TNO \cite{toet2017tno} (129 pairs), LLVIP \cite{Jia_Zhu_Li_Tang_Zhou_2021} (3463 pairs), and our constructed UMS (326 pairs). These datasets were synthetically generated to assess the generalization capability of our fusion network. Specially, the M\textsuperscript{3}FD, RoadScene, LLVIP, and TNO datasets lack semantic segmentation labels or object detection labels, which is not conducive to conducting experiments on downstream tasks. Therefore, we provide these two kinds of labels for the four datasets with YOLO-v7 \cite{Wang_Bochkovskiy_Liao} and the Segment-anything model \cite{Kirillov_Mintun_Ravi_Mao_Rolland_Gustafson_Xiao_Whitehead_Berg_Lo_etal}, which will be open-source in GitHub.

\textbf{Metrics:} structural similarity (SSIM), mean squared error (MSE), peak signal-to-noise ratio (PSNR), fidelity of visual information (VIF) \cite{Sheikh_Bovik_2004}, correlation coefficient (CC) \cite{Han_Tang_Gao_Hu_2013}, and Chen-Varshney metric (CV) \cite{Liu_Blasch_Xue_Zhao_Laganiere_Wu_2012} These metrics were chosen to capture different aspects of image quality, including structural integrity (SSIM, PSNR), pixel-level accuracy (MSE), visual fidelity (VIF), correlation between predicted and source image values (CC), and overall perceptual similarity (CV), thus providing a well-rounded assessment of image quality.

\textbf{Compared approaches:} Nine SOTA approaches were compare with the TSJNet: Reconet (Rec) \cite{Huang_Liu_Fan_Liu_Zhong_Luo}, CDDFuse (CDD) \cite{Zhao_Bai_Zhang_Zhang_Xu_Lin_Timofte_Gool_2022}, LRRNet (LRR) \cite{Li_Xu_Wu_Lu_Kittler_2023}, TarDal (TarD) \cite{Liu_Fan_Huang_Wu_Liu_Zhong_Luo}, IGNet (IGN) \cite{Li_Chen_Liu_Ma_2023}, SuperFusion (Sup) \cite{Tang_Deng_Ma_Huang_Ma_2022}, DIVFusion (DIV) \cite{Tang_Xiang_Zhang_Gong_Ma}, EMMA (EMM) \cite{07}, and SHIP(SHI) \cite{06}.
\subsection{Assessments of multi-modal image fusion}
\textbf{Subjective Comparisons.} In Fig.~\ref{04}, we present subjective outcomes for the MSRS, M\textsuperscript{3}FD, RoadScene, LLVIP, TNO, and UMS datasets. The images fused by TSJNet outperform those fused by the other SOTA methods. Specifically, TSJNet dramatically highlights targets, especially in low-light or overexposed areas, effectively distinguishing foreground targets from the background. Additionally, TSJNet retains rich edge and texture information that may be obscured in low-light conditions, such as car door frames, car front center nets, car logos, soldiers, bike wheels, and the trees.

\textbf{Objective Comparisons.} Table ~\ref{tab:comparison} lists the mean scores of six metrics for the six datasets. In general, TSJNet demonstrates superior objective performance across all six datasets, particularly excelling in SSIM, CC, and MSE, validating the high resemblance of fused results to the source images. The high PSNR values and low MSE values collectively reflect the stability of TSJNet in handling images from various scenes. TSJNet's emphasis on cross-modal feature extraction and preservation of edge details contributes to outstanding VIF values, indicating its high visual fidelity. Overall, the proposed method consistently extracts texture, brightness, and useful information from source images across different scenes while effectively capturing target information.
\subsection{Downstream applications}
\subsubsection{Object detection}
\textbf{Setup,} we performed object detection on M\textsuperscript{3}FD dataset, employed detector YOLO-v7 \cite{Wang_Bochkovskiy_Liao}, and evaluated the detection performance by the mean average precision (mAP) calculated at an intersection over union (IoU) of 0.5 metric (mAP@0.5). The training epoch, batch size,  and initial learning rate were set to 300, 16, and 0.01, respectively.

\textbf{Objective analysis.} As shown in Table~\ref{tab:seg/det comparisons}, the values of AP@0.5 and mAP@0.5 of the fusion results were higher than those of the unfused unimodal images. CDD, LRR, SHI and TSJNet exhibited good detection performances. Notably, the overall detection performance of TSJNet is 7.97\% higher than the average level of compared methods.

\begin{figure*}[htbp]
\centering	
\includegraphics[width=1.0\linewidth]{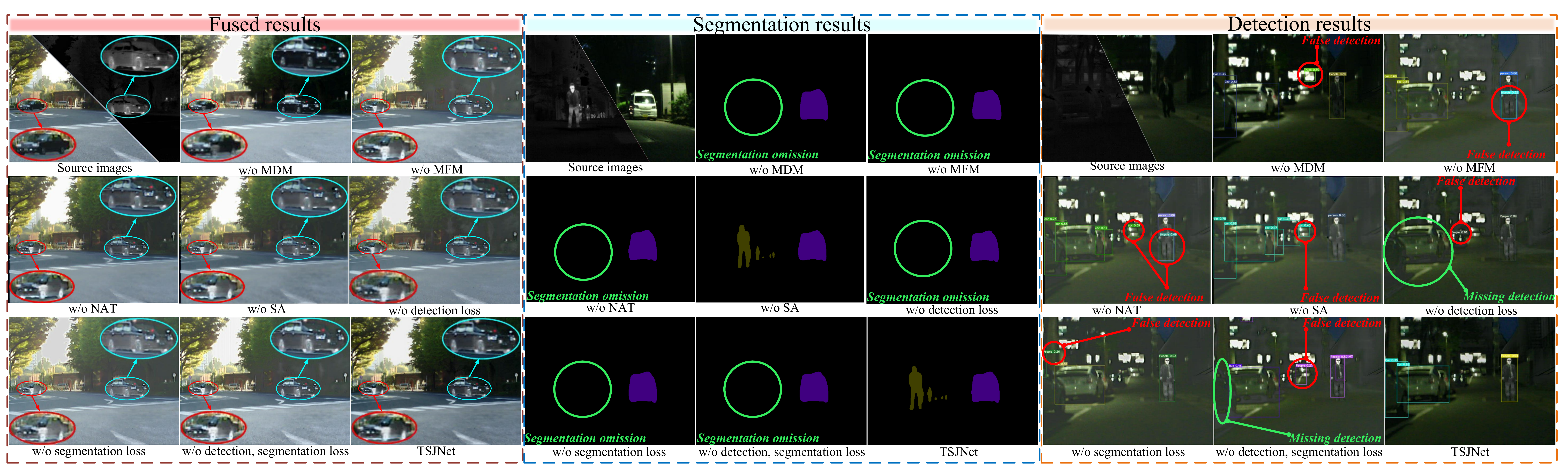}
\caption{Subjective ablation comparisons of the MDM, detection loss and segmentation loss.}
\vspace{-10pt}
\label{fig7}
\end{figure*}


\textbf{Subjective Analysis.} As shown in Fig.~\ref{object}, the detection results of M\textsuperscript{3}FD on images 04000 and 00275 are presented. The detection results of fused images, which incorporate complementary modal image features, generally outperform those of single-modal images. However, the impact of fusion performance on detection accuracy varies across different methods. In image 00275, ReC, CDD, LRR, TarD, and Sup mistakenly detect the reflection of a person in the rain as a real person. In image 04000, ReC, CDD, LRR, TarD, and IGN incorrectly classify a bucket as a person, while DIV exhibits low detection confidence. In contrast, the fusion results from TSJNet contain rich target features with high confidence, effectively avoiding these misclassifications.

\subsubsection{Semantic segmentation}
\textbf{Setup,} we used BiSeNet \cite{Yu_Wang_Peng_Gao_Yu_Sang_2018} to segment the semantic information of nine object classes of the MSRS dataset, \textit{i.e.}, background, bump, color cone, guardrail, curve, bike, person, car stop, and car. The model’s effectiveness was assessed by IoU. The number of training epochs was set to 200, the batch size was eight, and the remaining parameters remained unchanged, as in the initial experiment \cite{Yu_Wang_Peng_Gao_Yu_Sang_2018}.

\textbf{Subjective analysis.} Fig.~\ref{segmentation} shows the segmentation results for ``00939" and ``00770" of MSRS. In the left example, the ``yellow cones" cannot be segmented accurately by the compared methods, and ReC, LRR, and DIV cannot segment the distant cars. In the example on the right, ReC and TarD cannot segment any semantic information. CDD, IGN, Sup, DIV, EMM, and SHI have small errors in segmenting the ``people" at the edge position. The better segmentation of ``people" by TSJNet indicates that the fused images obtained by our model are better for semantic representation.

\textbf{Objective analysis.} Table~\ref{tab:seg/det comparisons} clearly demonstrates the segmentation superiority of TSJNet. The value of mIoU is the highest among the compared methods. Compared with its close competitors, CDD and Sup, TSJNet achieved higher scores in many important categories, such as curves and car stops. Furthermore, TSJNet exceeds the average of other methods by 10.88\% on the segmentation ability, which highlights the strong capabilities of our model in handling complex and detail-rich scenes.

\subsection{Ablation studies}
Ablation studies were conducted on the MSRS dataset. Table~\ref{tab:Ablation} summarizes the ablation results. 

\textbf{Ablation on the proposed MDM , MFM, and NAT.} The MDM, MFM, and NAT play a vital role in perceiving continuous information in the image background as well as salient texture of targets. From Fig.~\ref{fig7}, the image fused without the MDM has a limited ability for local brightness information extraction, \textit{e.g.}, the tires of a car. Moreover, the absence of either the MFM or NAT modules leads to the loss of fine structural details.

\textbf{Ablation on SA.} SA aims to enhance the model's adaptability and generalization ability to diverse scenarios. As illustrated in Fig.~\ref{fig7}, although the removal of the SA module exhibits minimal effect on segmentation performance, it tends to cause false detections in the object detection results.

\textbf{Ablation on detection and segmentation loss.} To verify the effectiveness of detection and segmentation double losses, we removed them separately and jointly from TSJNet, leaving the other parts unchanged. As shown in Table~\ref{tab:Ablation}, regardless of the type of loss removal, there is an overall downward trend in the metric values. As shown in Fig.~\ref{fig7}, rich semantic and target information are extracted using dual loss. Thus, images fused with cross-modal information concurrently can achieve both robust detection and precise segmentation results. It means that the employment of detection and segmentation losses can establish mutually advantageous relationships between fusion and high-level tasks.

\begin{table}[H]
\centering
\small
\caption{Results of ablation experiments on MSRS. }

\label{tab:Ablation}
\begin{tabular}{lccccc}
\toprule
Method & {SSIM$\uparrow$}&{PSNR$\uparrow$} & {MSE$\downarrow$} & {CC$\uparrow$} & {CV$\downarrow$} \\
\midrule
w/o MDM & 0.66 & 14.83 & 3731.88 & 0.54 & 560.46 \\
w/o MFM & 0.50 & 12.26 & 4218.04 & 0.46 & 451.98 \\
w/o NAT & 0.51 & 13.18 &3384.58  & 0.55 & 473.36 \\
w/o SA & 0.52 & 13.41& 3290.15 & 0.57 & 474.89 \\
w/o \(\mathcal{L}_{\text{Det}}\) & 0.55 & 14.86 & 2550.33 & 0.54 & 376.49 \\
w/o \(\mathcal{L}_{\text{Seg}}\) & 0.50 & 12.46 & 4167.93 &0.60 & 464.51 \\
w/o \(\mathcal{L}_{\text{Det, Seg}}\) & 0.49 & 12.43 & 4246.27 & 0.51 & 363.24 \\
\textbf{TSJNet} & \textcolor{top1}{0.69} & \textcolor{top1}{18.13} & \textcolor{top1}{1502.60} & \textcolor{top1}{0.62} & \textcolor{top1}{344.20} \\
\bottomrule
\end{tabular}
\vspace{-10pt} 
\end{table}

\section{Computational complexity analysis}

\begin{table}[htbp]
\centering
\caption{Comparison of model complexity, computation, and inference time}
\label{tab:5}
\begin{tabular}{lccc}
\toprule
\textbf{Method} & \textbf{Parameters (M)} & \textbf{FLOPs (G)} & \textbf{Time (s)} \\
\midrule
ReC     & $\textbf{0.008}$   & $\textbf{1.518}$ & 0.021 \\
CDD     & 1.186   & 29.213   & 0.363 \\
LRR     & 0.197   & 3.022    & 0.191 \\
TarD    & 0.297   & 91.138   & 0.084 \\
IGN     & 53.969  & 107.938  & $\textbf{0.014}$\\
Sup     & 1.962   & 76.630       & 0.070    \\
DIV     & 4.400   & 722.540  & 3.000 \\
EMM     & 1.520   & 17.724   & 0.065 \\
SHI     & 0.548   & 401.510  & 5.980 \\
TSJNet  & 45.070  & 2204.230 & 3.800 \\
\bottomrule
\end{tabular}
\vspace{-10pt} 
\end{table}

Table \ref{tab:5} compares the computational efficiency, FLOPs, and inference time of different methods. Although TSJNet does not demonstrate clear advantages in these aspects compared to other approaches, this is primarily due to its need to simultaneously optimize three tasks: image fusion, segmentation, and detection. By trading off higher model complexity and computational overhead, TSJNet achieves notable performance improvements in image fusion and its downstream tasks.

\section{Conclusion}
This paper proposes a novel multi-modal image fusion network termed TSJNet. By integrating an autoencoder architecture with a Multi-Dimensional Feature Extraction Module (MDM), the proposed method effectively fuses multi-scale shared semantic features and salient modality-specific features. Furthermore, the joint optimization of fusion, segmentation, and detection losses significantly enhances the representation of cross-modal semantic and target information. Meanwhile, a spatial attention module introduced at the decoding stage improves the model’s generalization capability across different data domains. In addition, we construct a UAV Multi-Scenario (UMS) benchmark dataset, which provides experimental support for the study and evaluation of multi-modal image fusion methods in real-world complex environments. Experimental results demonstrate that TSJNet can simultaneously enhance salient details and high-level semantic structures, thereby offering reliable support for downstream tasks such as object detection and semantic segmentation

Looking ahead, \rem{our future research} we will focus on improving the TSJNet to be more streamlined and highly efficient to meet the demands of multisource heterogeneous data fusion in practical contexts such as autonomous driving and drone military operations.

\section{Acknowledgment}
This research was supported by the National Natural Science Foundation of China(No. 62201149), the Basic and Applied Basic Research of Guangdong Province(No. 2023A1515140077), the Natural Science Foundation of Guangdong Province (No.2024A1515011880), and the Research Fund of Guangdong-HongKong-Macao Joint Laboratory for Intelligent Micro-Nano Optoelectronic Technology (No. 2020B1212030010).

 \bibliographystyle{elsarticle-num} 
 \bibliography{ref}






\end{document}